\documentclass[conference,unknownkeysallowed]{IEEEtran}

\usepackage{amsmath}          
\usepackage{amssymb} 			    
\usepackage{amsthm}           
\usepackage{textcomp}

\usepackage{float}            
\usepackage{cite}             
\usepackage{url}              

\usepackage{tabularx}         
\usepackage[table]{xcolor}    
\newcolumntype{Y}{>{\centering\arraybackslash}X}	

\usepackage{mathtools}          

\newcommand{\argmax}{\operatornamewithlimits{argmax}}

\usepackage{setspace}

\usepackage{booktabs} 

\usepackage{algorithm}
\usepackage[noend]{algpseudocode} 

\algnewcommand{\COMMENT}[2][.5\linewidth]{\leavevmode\hfill\makebox[#1][l]{//~#2}}
\algnewcommand{\LineComment}[1]{\State \(//\) #1}	
\algnewcommand\RETURN{\State \textbf{return} }

  \usepackage{graphicx}
  \usepackage[space]{grffile}   
  \graphicspath{{../figs/}}
  \DeclareGraphicsExtensions{.pdf,.jpeg,.png}

\usepackage[font={bf}]{caption}
\usepackage{paralist}
\setdefaultleftmargin{10pt}{10pt}{}{}{}{}

\usepackage{outlines}
\usepackage{enumitem}
\newcommand{\ItemSpacing}{0mm}
\newcommand{\ParSpacing}{0mm}
\setenumerate[1]{itemsep={\ItemSpacing},parsep={\ParSpacing},label=\arabic*.}
\setenumerate[2]{itemsep={\ItemSpacing},parsep={\ParSpacing}}

\usepackage[linewidth=1pt]{mdframed}
\mdfsetup{frametitlealignment=\center, skipabove=0, innertopmargin=1mm,
innerleftmargin=2mm,
innerrightmargin=2mm, leftmargin=0mm, rightmargin=0mm}

\newlength\myindent
\newtheorem{Definition}{Definition}[section]

\begin{document}


\title{EXTRACT: Strong Examples from \\ Weakly-Labeled Sensor Data}

\author{
\IEEEauthorblockN{Davis W. Blalock, John V. Guttag}
\IEEEauthorblockA{
Computer Science and Artificial Intellegence Laboratory \\
Massachussetts Institute of Technology \\
Cambridge, MA, USA \\
\{dblalock, guttag\}@mit.edu}
}

\maketitle

\begin{abstract}

Thanks to the rise of wearable and connected devices, sensor-generated time series comprise a large and growing fraction of the world's data. Unfortunately, extracting value from this data can be challenging, since sensors report low-level signals (e.g., acceleration), not the high-level events that are typically of interest (e.g., gestures). We introduce a technique to bridge this gap by automatically extracting examples of real-world events in low-level data, given only a rough estimate of when these events have taken place.

By identifying sets of features that repeat in the same temporal arrangement, we isolate examples of such diverse events as human actions, power consumption patterns, and spoken words with up to 96\% precision and recall. Our method is fast enough to run in real time and assumes only minimal knowledge of which variables are relevant or the lengths of events. Our evaluation uses numerous publicly available datasets and over 1 million samples of manually labeled sensor data.

\end{abstract}

\begin{IEEEkeywords}
Sensor data; Semi-supervised learning\footnote{This work to appear in IEEE International Conference on Data Mining 2016, \copyright IEEE 2016.}
\end{IEEEkeywords}

\section{Introduction} \label{sec:intro}

The rise of wearable technology and connected devices has made available a vast amount of sensor data, and with it the promise of improvements in everything from human health \cite{bsnSubdim} to user interfaces \cite{minnenAffixHMMs} to agriculture \cite{keoghInsect}. Unfortunately, the raw sequences of numbers comprising this data are often insufficient to offer value. For example, a smart watch user is not interested in their arm's acceleration signal, but rather in having their gestures or actions recognized.

Spotting such high-level events using low-level signals is challenging. Given enough labeled examples of the events taking place, one could, in principle, train a classifer for this purpose.
Unfortunately, obtaining labeled examples is an arduous task \cite{nuactiv, dataDicts, ushapelets, fastUshapelets}. While data such as images and text can be culled at scale from the internet, most time series data cannot. Furthermore, the uninterpretability of raw sequences of numbers often makes time series difficult or impossible for humans to annotate \cite{nuactiv}. 

It is often possible, however, to obtain \textit{approximate} labels for particular stretches of time. The widely-used human action dataset of \cite{msrc12}, for example, consists of streams of data in which a subject is known to have performed a particular action roughly a certain number of times, but the exact starts and ends of each action instance are unknown. Furthermore, the recordings include spans of time that do not correspond to any action instance. Similarly, the authors of the Gun-Point dataset obtained recordings \textit{containing} different gestures, but had to expend considerable effort extracting each instance \cite{ushapelets}. This issue of knowing that there are examples \textit{within} a time series but not knowing \textit{where} in the data they begin and end is common \cite{msrc12, ushapelets, dataDicts, minnenSubdim, nuactiv}. 

\begin{figure}[h]
\begin{center}
	\includegraphics[width=\linewidth]{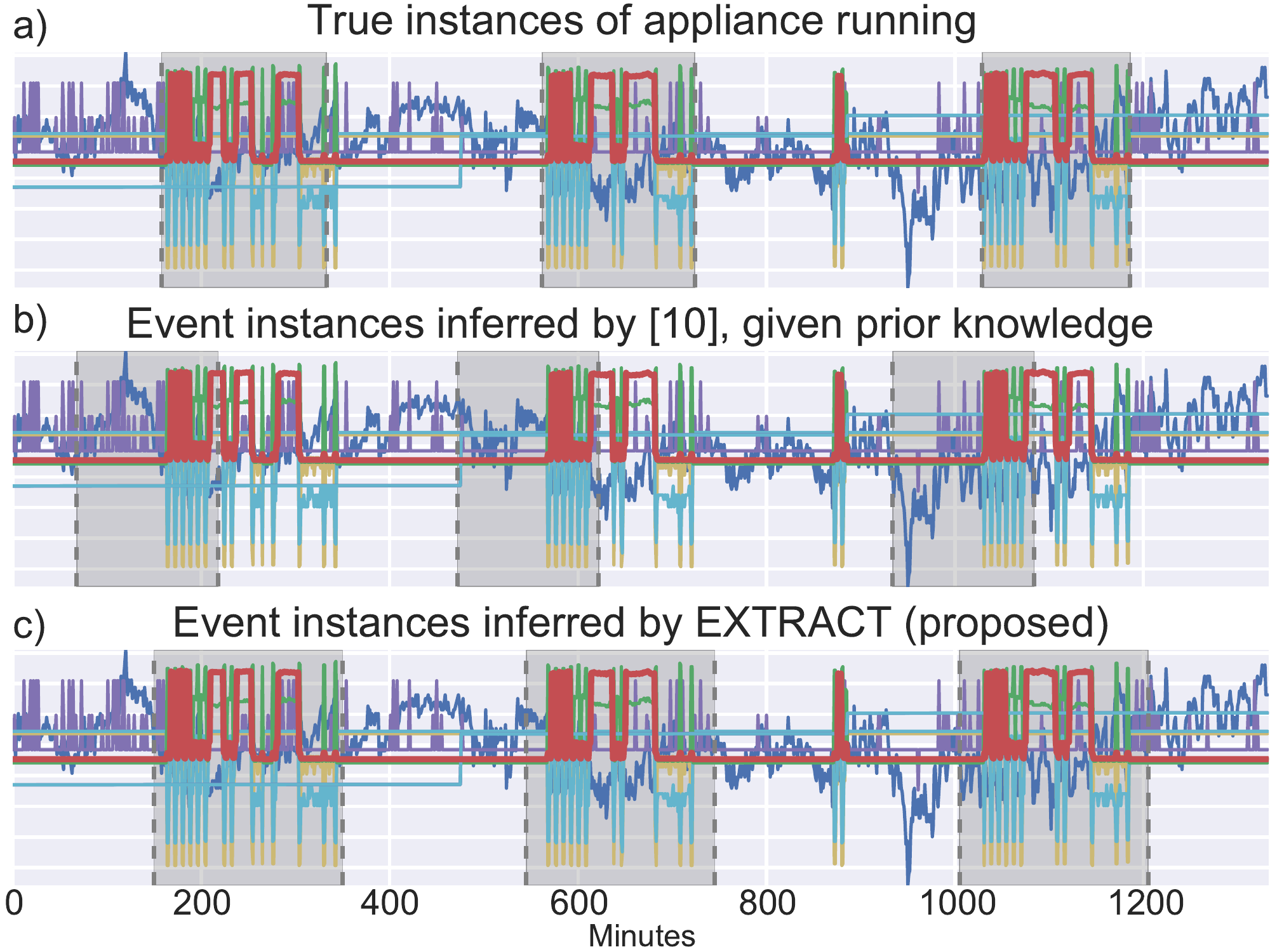}
	\vspace*{-4.5mm}
	\caption{\textit{a}) True instances of the dishwasher running (shaded). \textit{b}) Even when told the length and number of event instances, the recent algorithm of \cite{plmd} returns intervals with only the beginning of the event. \textit{c}) Our algorithm returns accurate intervals with no such prior knowledge.}
	\label{fig:fig1}
\end{center}
\end{figure}

To leverage these \textit{weak labels}, we developed an algorithm, EXTRACT, that efficiently isolates examples of an event given only a time series known to contain several occurrences of it. A simple illustration of the problem we consider is given in Figure~\ref{fig:fig1}. The various lines depict the (normalized) current, voltage, and other power measures of a home dishwasher. Shown are three instances of the dishwasher running, with idleness in between. With no prior knowledge or domain-specific tuning, our algorithm correctly determines not only what this repeating event looks like, but also where it begins and ends.
\vspace{-.5mm}

\vspace{-1mm}
This is a challenging task, since the variables affected by the event, as well as the number, lengths, and positions of event instances, are all unknowns. Further, it is not even clear what objective should be maximized to find an event. For example, finding the nearest subsequences using the Euclidean distance yields the incorrect event boundaries returned by \cite{plmd} (Fig~\ref{fig:fig1}b).

To overcome these barriers, our technique leverages three observations:
\vspace{-.25mm}
\begin{enumerate}
\itemsep0em
\item Each subsequence of a time series can be seen as having representative features; for example, it may resemble different shapelets \cite{shapelets} or have a particular level of variance.
\item A repeating event will cause a disproportionate number of these features to occur together where the event happens. In Figure~\ref{fig:fig1}, for example, these features are a characteristic arrangement of spikes in the values of certain variables.
\item If we can identify these features, we can locate each instance of the event with high probability. This holds even in the presence of irrelevant variables (which merely fail to contribute useful features) and unknown instance lengths (which can be inferred based on the interval over which the features occur together).
\end{enumerate}

\hspace{-10pt}Our contributions consist of:
\vspace{-.25mm}
\begin{itemize}
\itemsep-.1mm
\item A formulation of semi-supervised event discovery in time series under assumptions consistent with real data. In particular, we allow multivariate time series, events that affect only subsets of variables, and instances of varying lengths.
\item An $O(N\log(N))$ algorithm to discover event instances under this formulation. It requires less than 300 lines of code and is fast enough to run on batches of data in real time. It is also considerably faster, and often much more accurate, than similar existing algorithms \cite{plmd, moen}. For example, we recognize instances of the above dishwasher pattern with an F1 score of over $90$\%, while the best-performing comparison \cite{plmd} achieves under $30$\%.
\item Open source code and labeled time series that can be used to reproduce and extend our work. In particular, we believe that our annotation of the full dishwasher dataset \cite{ampds} makes this the longest available sensor-generated time series with ground truth event start and end times.
\end{itemize}


\vspace{-1mm}
\section{Definitions and Problem} \label{sec:problem}

\begin{Definition}{
\textbf{Time Series}. A $D$-dimensional time series $T$ of length $N$ is a sequence of real-valued vectors $t_1,\ldots,t_N, t_i \in \mathbb{R}^D$. If $D = 1$, we call $T$ ``univariate'' or ``one-dimensional,'' and if $D > 1$, we call it ``multivariate'' or ``multi-dimensional.''
}
\vspace{-2mm}
\end{Definition}
\begin{Definition}{\textbf{Region}. A region $R$ is a pair of indices $(a, b), a \le b$. The value $b - a + 1$ is termed the \textbf{length} of the region, and the time series $t_a,\ldots,t_b$ is the \textbf{subsequence} for that region. If a region reflects an occurrence of the event, we term the region an event \textbf{instance}.
}
\end{Definition}

\subsection{Problem Statement}
We seek the set of regions that are most likely to have come from a shared ``event'' distribution rather than a ``noise'' distribution. This likelihood is assessed based on the subset of features maximizing how distinct these distributions are (using some fixed feature representation).

Formally, let $x_1,\ldots,x_K$ be binary feature representations of all $K$ possible regions in a given time series and $x_i^j$ denote feature $j$ in the region $i$. We seek the optimal set of regions $\mathcal{R}^{\ast}$, defined as:
\begin{align} \label{eq:concreteObjective}
	\mathcal{R}^{\ast}= \argmax_{\mathcal{R}} \max_{\mathcal{F}} p(\mathcal{R}) \sum_{j \in \mathcal{F}} c_j (log(\theta_{1j}) - log(\theta_{0j}))
\end{align}
where $\theta_{0j}$ and $\theta_{1j}$ are the empirical probabilities for each feature $j$ in the whole time series and the regions $\mathcal{R}$ respectively, and $c_j$ is the count of feature $j$, ${\sum_{i \in \mathcal{R}} x_i^j}$. $\mathcal{F}$ is the set of features that best separate the event. The prior $p(\mathcal{R})$ is 0 if regions overlap too heavily or violate certain length bounds (see below) and is otherwise uniform.

Equation~\ref{eq:concreteObjective} says that we would like to find regions $i$ and features $j$ such that $x_i^j$ happens both many times (so that $c_j$ is large) and much more often than would occur by chance (so that $log(\theta_{1j}) - log(\theta_{0j})$ is large). In other words, the best features $\mathcal{F}$ are the largest set that consistently occurs across the most regions, and $\mathcal{R}^{\ast}$ is these regions.

Given certain independencies, this objective is a MAP estimate of the regions and features. Because of space constraints, we defer the details to \cite{extractWebsite}.

\vspace{-1mm}
\subsection{Assumptions}
\vspace{-.5mm}
\hspace{-10pt}We \textit{do not} make any of the following common assumptions: 
\vspace{-.5mm}
\begin{itemize}
\itemsep.2mm
\item A known number of instances. 
\item A known or constant length for instances.
\item A known or regular spacing between instances.
\item A known set of characteristics shared by instances. In particular, we do not assume that all instances have the same mean and variance, so we cannot bypass normalization when making similarity comparisons.
\item That there is only one dimension.
\item That all dimensions are affected by the event.
\item Anything about dimensions not affected by the event.
\end{itemize}

\hspace{-10pt}So that the problem is well-defined, we \textit{do} assume that:
\begin{itemize}
\itemsep.2mm
\item The time series contains instances of only one class of event. It may contain other transient phenomena, but we take our weak label to mean (only) that the primary structure in the time series comes from the events of the labeled class and that there are no other repeating events.
\item There are at least two instances of the event, and each instance produces some characteristic (but unknown) pattern in the data.
\item There exist bounds $M_{min}$ and $M_{max}$, $M_{min} > M_{max}/2$ on the lengths of instances. These bounds disambiguate the case where pairs of adjacent instances could be viewed as single instances of a longer event. Similarly, no two instances overlap by more than ${M_{min}-1}$ time steps.
\end{itemize}
\vspace{-1mm}
We also do not consider datasets in which instances are rare \cite{rareMotif}---all time series used in our experiments have instances that collectively comprise $\sim$10\% of the data or more (though this exact number is not significant).

\subsection{Why the Task is Difficult}

The lack of assumptions means that the number of possible sets of regions and relevant dimensions is intractably large. Suppose that we have a $D$-dimensional time series $T$ of length $N$ and $M_{min} \le M \le M_{max}$. There are up to $O(N/M)$ instances, which can collectively start at (at most) $\binom{N}{N/M}$ positions. Further, each can be of $O(M)$ different lengths. Finally, the event may affect any of $2^D - 1$ possible subsets of dimensions. Altogether, this means that there are roughly $O(N^{N/M} \cdot M^{N/M} \cdot 2^D)$ combinations of regions and dimensions. 

Moreover, while there may be heuristics or engineered features that could allow isolation of any particular event in any particular domain, we seek to develop a general-purpose tool that requires no coding or tuning by humans. We therefore do not use such event-specific knowledge. This generality is both a convenience for human practitioners and a necessity for real-world deployment of a system that learns new events at runtime. 


Lastly, because our aim is to extract examples for future use, we seek to locate full events, not merely the pieces that are easiest to find.

\vspace{-1mm}
\section{Related Work} \label{sec:relatedWork}

Several authors have built algorithms to address the difficulty of obtaining labeled time series for various tasks. The authors of \cite{ushapelets} and \cite{fastUshapelets} cluster univariate time series when much of the data in each time series is irrelevant. They do this by discovering informative shapelets \cite{shapelets} in an unsupervised manner. Their goal is to assign entire time series to various clusters. In contrast, we are interested in assigning a subset of the regions within a single time series to a particular ``cluster.''

The Data Dictionaries of \cite{dataDicts} are closer to sharing our problem formulation in that they too find class-related subsequences within a weakly-labeled time series. However, they are interested in frame-level, rather than event-level, classification. They also assume a user-specified query length, that all classes are known, and that all variables are relevant.

Methodologically, the algorithms of \cite{unsupervisedMusicBoundaries} and \cite{crossmatch} are similar to our own. However, the former technique assumes all regions of a time series reflect various ongoing phenomenon, and the latter relies on instances sharing a common mean and variance. In terms of representation, the dot plots of \cite{keoghDotPlots} are similar to our work, but the authors use them only for human inspection, rather than algorithmic mining. They also require the setting of multiple user-specified parameters. 


There is also a vast body of work on unsupervised discovery of repeating patterns in time series, typically termed ``motif discovery.'' Most of these works consider univariate time series and/or the task of finding only the closest pair of regions under some distance measure \cite{keoghUSMotifs, motifOrig}. Others consider the task of finding multiple motifs and/or refining motif results produced by other algorithms \cite{gstex, minnenAffixHMMs, minnenSubdim, moen, plmd}, both of which are orthogonal to our work in that they could employ our algorithm as the basic motif-finding subroutine.

A few motif discovery works seek to find all instances of a given event as we do, albeit under different assumptions. The techiques of \cite{plmd}, \cite{ratanaFindAllCrap}, and \cite{epenthesis} do so by finding closest pairs of subsequences at different lengths and then extracting subsequences that are sufficiently similar under an entropy-based measure. Those of \cite{moen} and \cite{minnenSubdim} do much the same, although with a distance-based generalization heuristic. All except \cite{epenthesis} assume that event instances share a single length, and all but \cite{minnenSubdim} assume that all dimensions are relevant. We discuss \cite{plmd}, \cite{moen}, and \cite{epenthesis} further in Section~\ref{sec:results}.


\vspace{-1mm}
\section{Method Overview} \label{sec:intuition}

In this section we offer a high-level overview of our technique and the intuition behind it, deferring details to Section~\ref{sec:method}.

The basic steps are given in Algorithm~\ref{algo:extract}. In step 1, we create a representation of the time series that is invariant to instance length and enables rapid pruning of irrelevant dimensions. In step 2, we find sets of regions that may contain event instances. In step 3, we refine these sets to estimate the optimal instances $\mathcal{R}^\ast$.

Since the main challenge overcome by this technique is the lack of information regarding instance lengths, instance start positions, and relevant dimensions, we elaborate upon these steps by describing how they allow us to deal with each of these unknowns. We begin with a simplified approach and build to a sketch of the full algorithm. In particular, we begin by assuming that time series are one-dimensional, instances are nearly identical, and all features are useful.

\begin{algorithm}[h] 
\caption{EXTRACT}
\label{algo:extract}
\end{algorithm}
\vspace{-3mm}
\begin{mdframed}
\begin{outline}[enumerate]
\vspace{-2mm}
\1 \hspace{-1.5mm} \textbf{Transform the time series $T$ into a feature matrix $\Phi$}
\vspace{-.5mm}
	\2 Sample subsequences from the time series to use as shape features
	\2 Transform $T$ into $\Phi$ by encoding the presence of these shapes across time
	\2 Blur $\Phi$ to achieve length and time warping invariance

\1 \textbf{Using $\Phi$, generate sets of ``candidate'' windows that may contain event instances}
\vspace{-.5mm}
	\2 Find ``seed'' windows that are unlikely to have arisen by chance
	\2 Find ``candidate'' windows that resemble each seed
	\2 Rank candidates based on similarity to their seeds

\1 \textbf{Infer the true instances within these candidates}
\vspace{-.5mm}
	\2 Greedily construct subsets of candidate windows based on ranks
	\2 Score these subsets and select the best one
	\2 Infer exact instance boundaries within the selected windows
\end{outline}
\end{mdframed}

\subsection{Unknown Instance Lengths}
\begin{figure}[h]
\begin{center}
	\includegraphics[width=\linewidth]{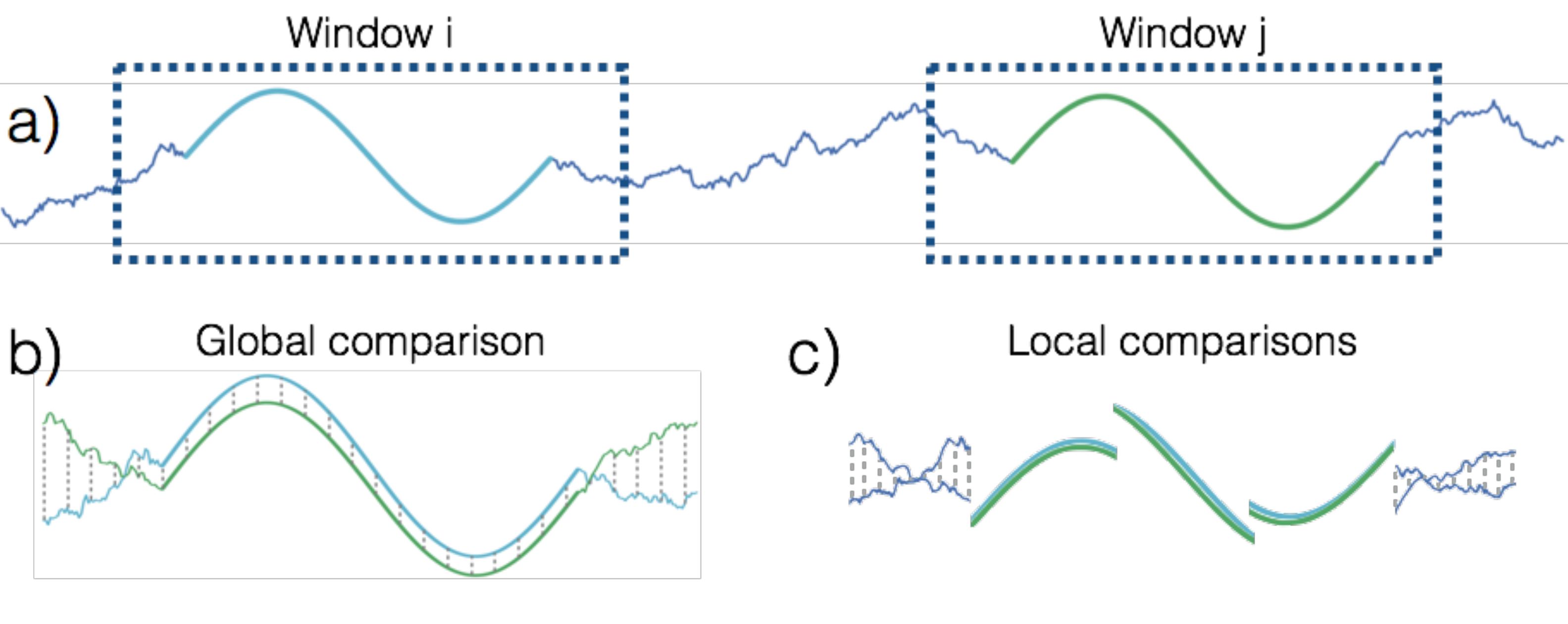}
	\caption{a) A time series containing two event instances. b) Including extra data yields large distances (grey lines) between the windows $i$ and $j$ around the event instances. c) Comparing based on subsequences within these windows allows close matches between the pieces of the sine waves.}
	\label{fig:lengthsProblem}
\end{center}
\end{figure}

Like most existing work, we find event instances by searching over shorter regions of data within the overall time series (Fig \ref{fig:lengthsProblem}a). Since we do not know how long the instances are, this seemingly requires exhaustively searching regions of many lengths \cite{moen, plmd, ratanaFindAllCrap}, so that the instances are sure to be included in the search space.

However, there is an alternative. Our approach is to search over all regions of a \textit{single} length $M_{max}$ (the maximum possible instance length) and then refine these approximate regions. For the moment, we defer details of the refinement process. We refer to this set of regions searched as \textit{windows}, since they correspond to all positions of a sliding window over the time series.

This single-length approach presents a challenge: since windows longer than the instances will contain extraneous data, only parts of these windows will appear similar. As an example, consider Figure~\ref{fig:lengthsProblem}. Although the two windows shown contain identical sine waves, the noise around them causes the windows to appear different, as measured by Euclidean distance (the standard measure in most motif discovery work) (Fig~\ref{fig:lengthsProblem}b). Worse, because the data must be mean-normalized for this comparison to be meaningful \cite{mk}, it is not even clear what portions of the regions are similar or different---because the noise has altered the mean, even the would-be identical portions are offset from one another.
\vspace{-1mm}

However, while the windows appear different when treated as atomic objects, they have many sub-regions (namely, pieces of the sine waves) that are similar when considered in isolation (Fig \ref{fig:lengthsProblem}c). This suggests that if we were to compare the windows based on \textit{local} characteristics, instead of their \textit{global} shape, we could search at a length longer than the event and still determine that windows containing event instances were similar.

\begin{figure}[h]
\begin{center}
	\includegraphics[width=\linewidth]{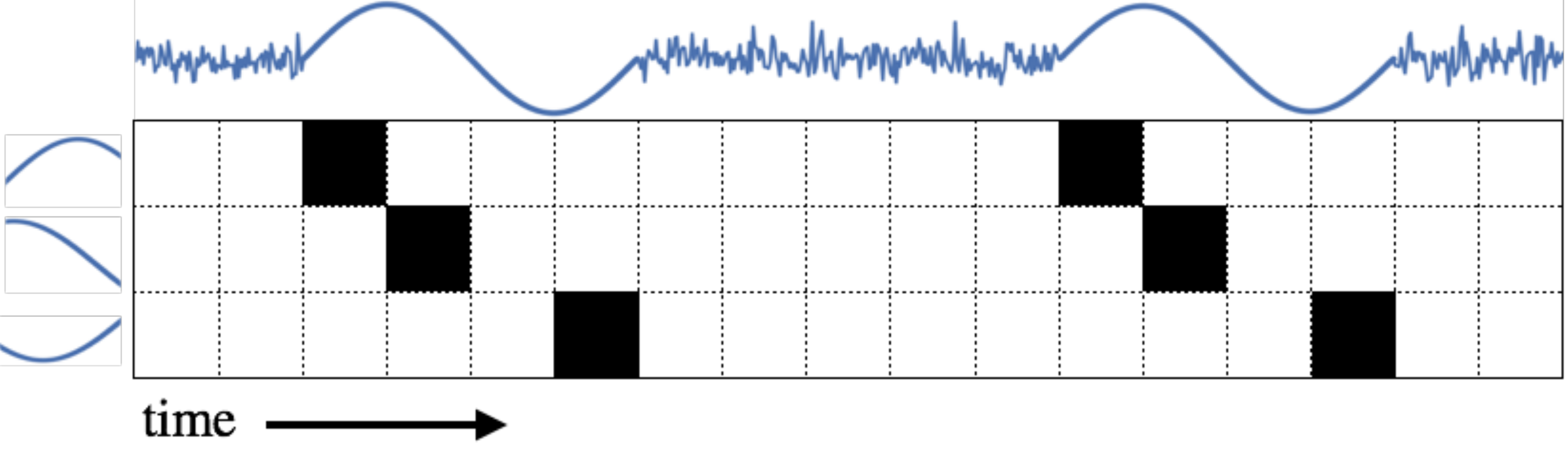}
	\caption{Feature matrix. Each row is the presence of a particular shape, and each column is a particular time (shown at reduced granularity). Similar regions of data contain the same shapes in roughly the same temporal arrangement.}
	\label{fig:basicTransform}
\end{center}
\end{figure}

To enable this, we transform the data into a sparse binary feature matrix that encodes the presence of particular shapes at each position in the time series (Fig \ref{fig:basicTransform}). Columns of the feature matrix are shown at a coarse granularity for visual clarity--in reality, there is one column per time step. We defer explanation of how these shapes are selected and how this feature matrix is constructed to the next section.

\begin{figure}[h]
\begin{center}
	\includegraphics[width=\linewidth]{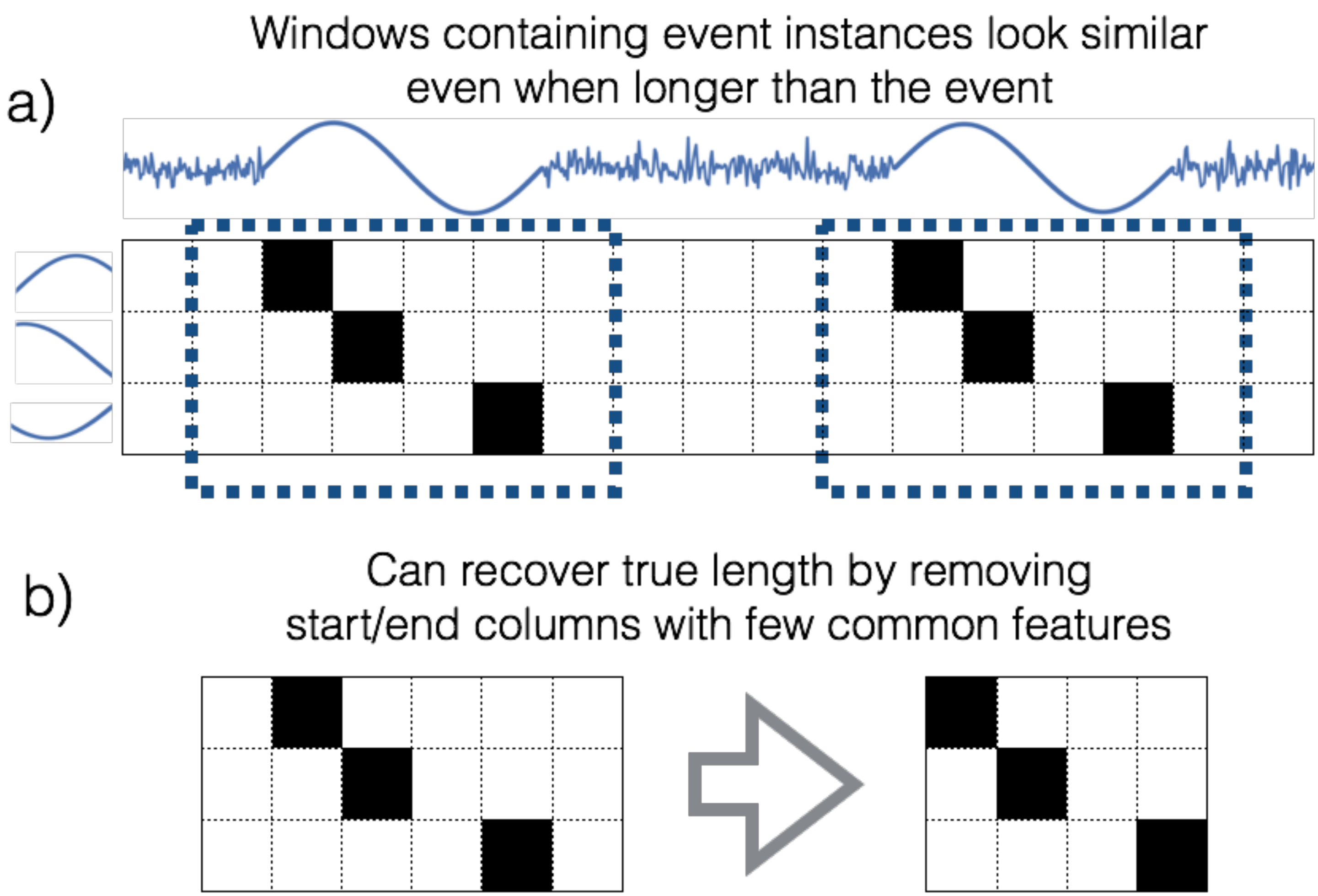}
	\caption{Because the values in the feature matrix are independent of the window length, a window longer than the event can be used to search for instances.}
	\label{fig:windowTooLong}
\end{center}
\end{figure}

Using this feature matrix, we can compare windows of data without knowing the lengths of instances. This is because, even if there is extraneous data at the ends of the windows, there will still be more common features where the event happens (Fig \ref{fig:windowTooLong}a) than would be expected by chance.

Once we identify the windows containing instances, we can recover the starts and ends of the instances by examining which columns in the corresponding windows look sufficiently \\ similar---if a start or end column does not contain a consistent set of 1s across these windows, it is probably not part of the event, and we prune it (Fig \ref{fig:windowTooLong}b).

Unfortunately, this figure is optimistic about the regularity of shapes within instances. In reality, a given shape will not necessarily be present in all instances, and a set of shapes may not appear in precisely the same temporal arrangement more than once because of uniform scaling \cite{keoghUSMotifs} and time warping. We defer treatment of the first point to a later section, but the second can be remedied with a preprocessing step.
\vspace{-.5mm}

\begin{figure}[h]
\begin{center}
	\includegraphics[width=\linewidth]{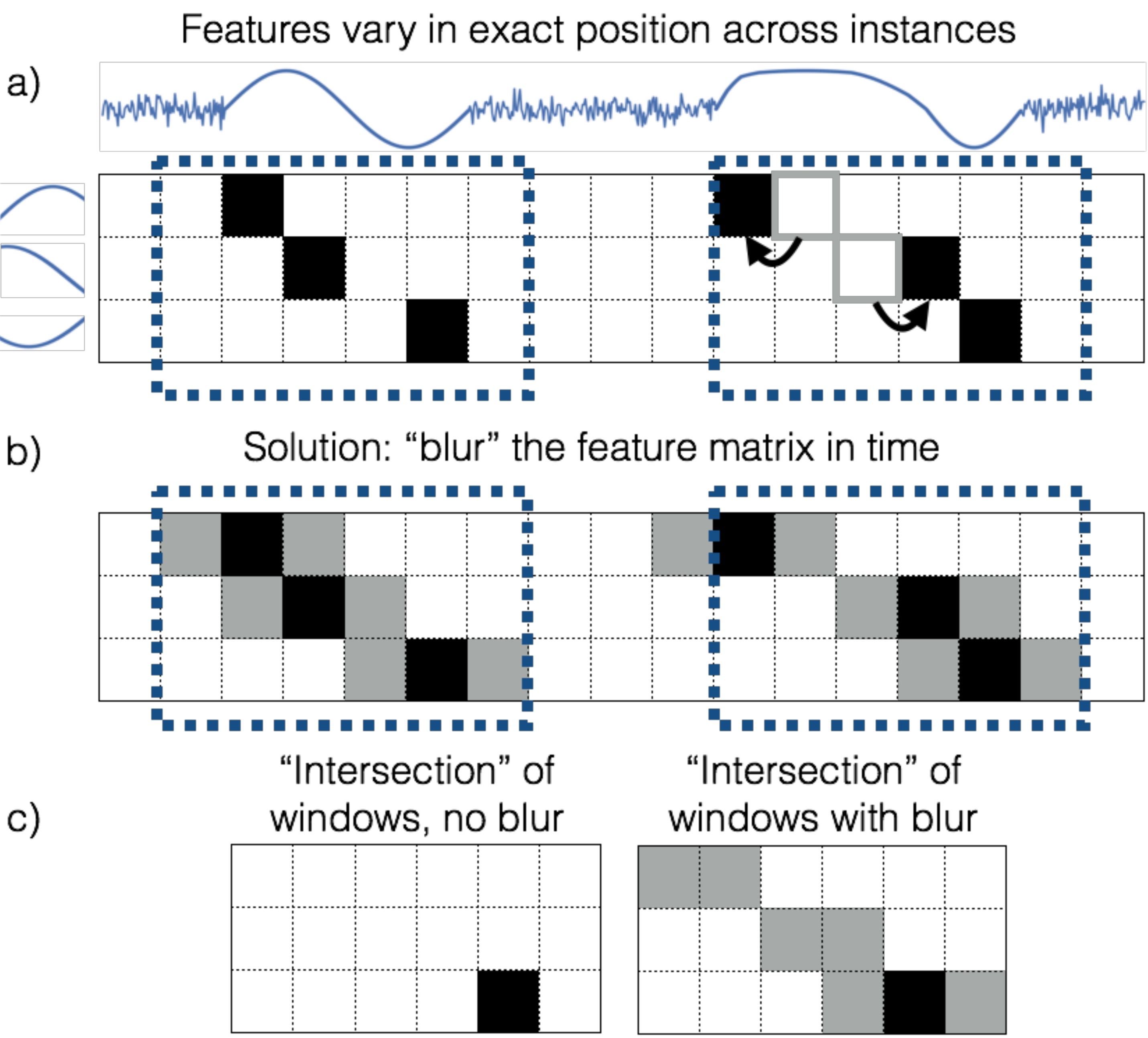}
	\caption{Blurring the feature matrix. Despite the second sine wave being longer and warped in time, the two windows still appear similar when blurred.}
	\label{fig:blur}
\end{center}
\end{figure}

To handle both uniform scaling and time warping simultaneously, we ``blur'' the feature matrix in time. The effect is that a given shape is counted as being present over an interval, rather than at a single time step. This is shown in Figure \ref{fig:blur}, using the intersection of the features in two windows as a simplified illustration of how similar they are. Since the blurred features are no longer binary, we depict the ``intersection'' as the elementwise minimum of the windows.
\vspace{-3mm}

\vspace{2mm}
\subsection{Dealing with Irrelevant Features}

Thus far, we have assumed that the shapes encoded in the matrix are all characteristic of the event. In reality, we do not know ahead of time which shapes are relevant, and so there will also be many irrelevant features.

\begin{figure}[h]
\begin{center}
	\includegraphics[width=\linewidth]{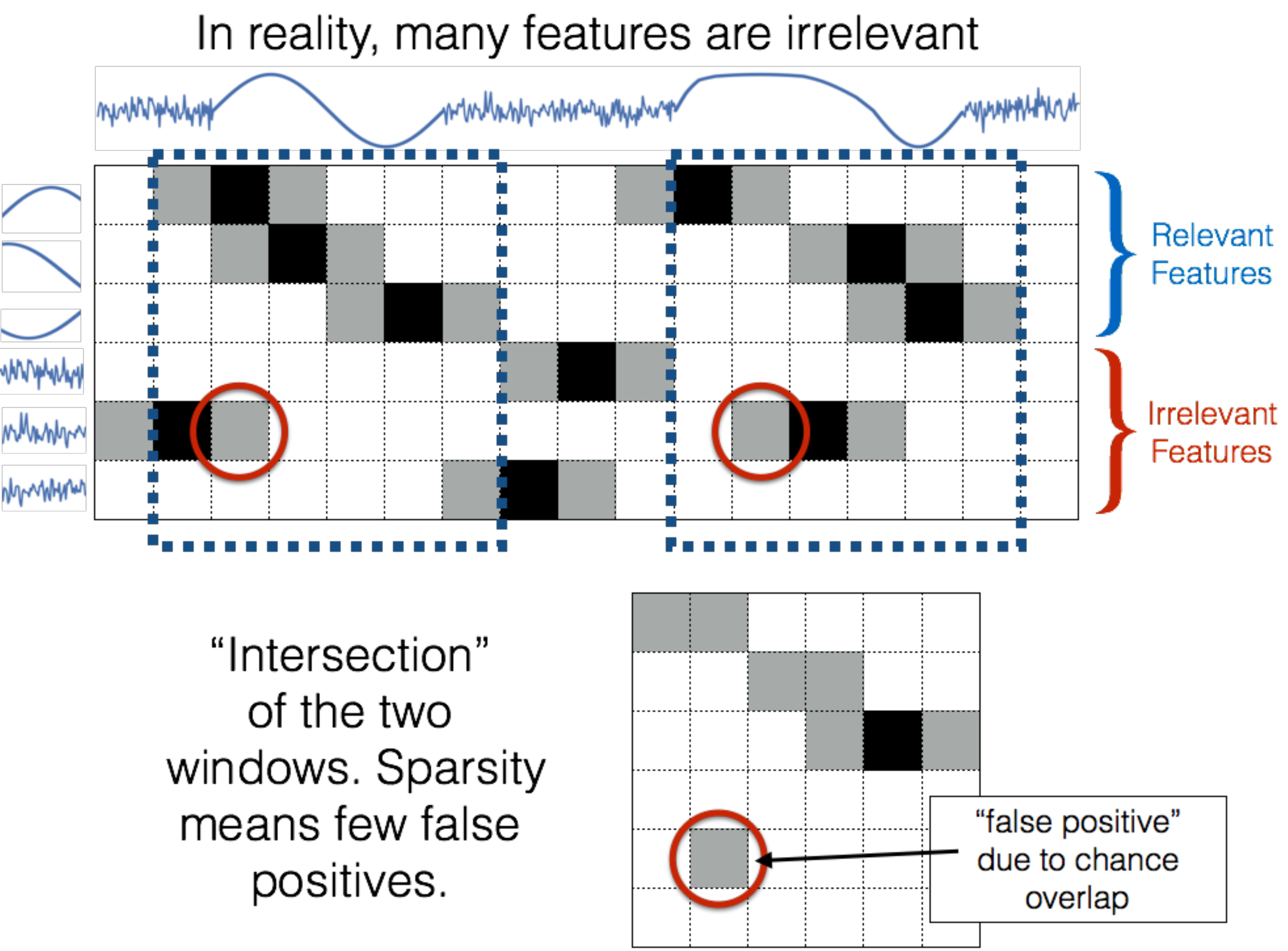}
	\caption{Most irrelevant features are ignored after only two or three examples, since it is unlikely for them to repeatedly be in the same place in the window. A few ``false positives'' may remain.}
	\label{fig:irrelevantFeatures}
\end{center}
\end{figure}

Fortunately, the combination of sparsity and our ``intersection'' operation causes us to ignore these extra features (Fig~\ref{fig:irrelevantFeatures}). To see this, suppose that the probability of an irrelevant feature being present at a particular location in an instance-containing window is $p_0$. Then the probability of it being present by chance in $k$ windows is $\approx p_0^k$. Feature matrices for real-world data are over $90$\% zeros, since a given subsequence can only resemble a few shapes. Consequently, $p_0$ is small (e.g., $0.05$), and $p_0^k \approx 0$ even for small $k$.

\vspace{-1mm}
\subsection{Multiple Dimensions}
\vspace{-1mm}

The generalization to multiple dimensions is straightforward: we construct a feature matrix for each dimension and concatenate them row-wise. That is, we take the union of the features from each dimension. A dimension may not be relevant, but this just means that it will add irrelevant features. Thanks to the aforementioned combination of sparsity and the intersection operation, we ignore these features with high probability.

\subsection{Finding Instances} \label{sec:basicAlgo}
The previous subsections have described how we construct the feature matrix. In this section, we describe how to use this matrix to find event instances. A summary is given in Algorithm~\ref{algo:findInstancesShort}. The idea is that if we are given one ``seed'' window that contains an instance, we can generate a set of similar ``candidate'' windows and then determine which of these are likely event instances. Since we cannot generate seeds that are certain to contain instances, we generate many seeds and try each. We defer explanation of how seeds are generated to the next section.

\begin{algorithm}[h]
\caption{$FindInstances(\mathcal{S}, \mathcal{X})$ }
\label{algo:findInstancesShort}
\begin{algorithmic}[1]

\State \textbf{Input:} $S$, ``seed'' windows; $\mathcal{X}$, all blurred windows

\State $\mathcal{I}_{best} \leftarrow \{\}$
\State $score_{best} \leftarrow -\infty$
\For {each seed window $s \in \mathcal{S}$}
	\LineComment{Find candidate windows $\mathcal{C}$ based on}
	\LineComment{dot product with seed window $s$}
	\State $P \leftarrow [s \cdot \tilde{x}_i, \tilde{x}_i \in \mathcal{X}]$
	\State $\mathcal{C} \leftarrow localMaxima(P)$
	\State $\mathcal{C} \leftarrow enforceMinimumSpacing(\mathcal{C}, M_{min})$
	\State $\mathcal{C} \leftarrow sortByDescendingDotProd(\mathcal{C}, P)$
	\LineComment{assess subsets of $\mathcal{C}$}
	\For {$k \leftarrow 2,\ldots,|\mathcal{C}|$}
		\State $\mathcal{I} \leftarrow \{c_1,...,c_k\}$ \COMMENT{$k$ best candidates}
		\State $score \leftarrow computeScore(\mathcal{I}, c_{k+1})$
		\If { $score > score_{best}$ }
			\State $score_{best} \leftarrow score$
			\State $\mathcal{I}_{best} \leftarrow \mathcal{I}$
		\EndIf
	\EndFor
\EndFor
\Return $\mathcal{I}_{best}$
\vspace{-1mm}
\end{algorithmic}
\end{algorithm}

The main loop iterates through all seeds $s$ and generates sets of candidate windows for each. These candidates are the windows whose dot products with $s$ are local maxima---i.e., they are higher than those of the windows just before and after. To prevent excess overlap, a minimum spacing is enforced between the candidates by only taking the best relative maximum in any interval of width $M_{min}$ (the instance length lower bound). If $s$ contains an event instance, the resulting candidates should be (and typically are) a superset of the true instance-containing windows.

In the inner loop, we assess subsets of the candidates to determine which ones contain instances. Since there are $2^{|\mathcal{C}|} = O(2^{N/M_{min}})$ possible subsets, we use a greedy approach that tries only $|\mathcal{C}| = O(N/M_{min})$ subsets. Specifically, we rank the candidates based on their dot products with $s$ and assess subsets that contain the $k$ highest-ranking candidates for each possible $k$.

The final set returned is the highest-scoring subset of candidates for any seed. See Section~\ref{sec:scoring} for an explanation of the scoring function.

\vspace{-1mm}
\section{Method Details} \label{sec:method}

We now describe how the ideas of the previous section translate into a concrete algorithm.
Throughout this section, let $T$ denote a $D$-dimensional time series of length $N$, $M_{min}$ and $M_{max}$ denote the instance length bounds, $S$ denote the set of seed indices, $\Phi = \{\phi_1,\ldots,\phi_N\}$, $\phi_i \in \mathbb{R}^J$ denote the feature matrix, and $X = \{x_1,\ldots,x_{N-M_{max}+1}\}$ denote the data in $\Phi$ for each possible sliding window position; i.e., $x_i$ = $\Phi_{:,i:i+M_{max}}$. Further, let $\widetilde{\Phi}$ denote the blurred feature matrix and $\widetilde{X}$ denote the windows of data in $\widetilde{\Phi}$.













\subsection{Structure Scores}
We select shape features and seed regions that appear most likely to have been generated by some latent event. Since we lack domain-specific knowledge about what distinguishes such regions, we use the common approach of modeling ``non-event'' time series as random walks \cite{rareMotif, plmd}. Specifically, let $T$ be a univariate time series of length $N$, and $W_1,\ldots,W_{100}$ be a collection of 100 Gaussian random walks\footnote{The exact number of walks is unimportant; using larger values (e.g., 1000 or 10,000) has no effect.} of length $N$ with $\sigma^2 = E[(t_i - t_{i-1})^2]$. We define:
\vspace{-2mm}
\begin{align}
	structure(T) = \min_{W} \frac{1}{N} \sum_{i=1}^{N} ((t_i - \mu_T) - (w_i - \mu_W))^2
\end{align}
where $\mu_T$ and $\mu_W$ are the means of the time series. The score is the minimum squared Euclidean distance to any of the random walks, normalized by mean and length. For multivariate time series, we sum the scores for each dimension. This is an approximation to the negative log likelihood of $T$ being a random walk, using the optimal $\sigma^2$.

\subsection{Constructing the Feature Matrix} \label{sec:featureMat}

The first step in building the feature matrix is selecting the lengths of the shapes to use as features. Since we do not know what length is best, we employ all lengths that are powers of two within the interval $[8, M_{max}]$ samples. $8$ is used because $2$ or $4$ samples are not enough to define a meaningful shape.

For each length $M$ and each dimension, we select shapes by randomly sampling from the data. To limit the algorithm's complexity to $O(N\log(N))$, we select $\log(N)$ subsequences. The probability of each subsequence being selected is proportional to its structure score.

For each shape $j$, we construct its row in the feature matrix $\Phi_{j}$ by sliding it over the data in its dimension and setting the value to 1 iff the distance between the shape and the subsequence centered at each position is less than some threshold. This threshold is fixed at 0.25 since this robustly rejects random walk data and consistently worked better than .125 or .5 in preliminary experiments.\footnote{On small sets of time series not used in the reported experiments. Note too that an all-zero subsequence yields a distance of 1, so the threshold must be a number below this.} If a shape has no subsequences within the threshold (other than itself), its row is discarded to save computation. The row is also discarded if it is more than half nonzeros, which happens in constant signals. Distances are given by the mean-normalized Euclidean distance squared between the shape and subsequence, divided by their length and the variance of the shape.

To construct the blurred feature matrix $\widetilde{\Phi}$, we convolve each row with a Hamming filter of length $M_{min}$. We then divide each entry by the largest value within $M_{min} / 2$ time steps in its row, so that the maximum value in $\widetilde{\Phi}$ remains 1.

\vspace{-1mm}
\subsection{Generating Seed Windows}

We generate seeds by finding the start indices associated with the highest structure scores. Concretely, we score each start index $t$ as the sum of the structure scores of the subsequences of all power-of-two lengths $M \in [8, M_{min}]$ that begin at $t$. We then take the two best start indices at least $M_{min}$ apart. One could use any constant number of seeds without affecting the complexity, and we choose two because using more has little or no impact on accuracy. Since these two seeds are unlikely to be exact instance starts, we add 10 additional seeds on either side of each, spaced $M_{max} / 10$ apart.\footnote{If one of the initial two seeds is within $M_{max}$ of an instance start, adding $\alpha$ more seeds $\frac{M_{max}}{\alpha}$ apart on either side guarantees that one of them will be within $\frac{M_{max}}{2\alpha}$ of an instance start. $\alpha = 10$ is an arbitrary value large enough to avoid overly-spaced seeds missing the start as a source of error.} 

This seed generation scheme is a heuristic, but we found that it worked better in practice than other heuristics. For example, using the two indices with the best structure scores yielded higher accuracy than using the indices of the closest pair of subsequences under the z-normalized Euclidean distance, as in \cite{moen, plmd, ratanaFindAllCrap}. If the data contained many subsequences that appeared to be non-random but were not instances, a different heuristic would be required. One could also supply a single known instance start as the lone seed to bypass the need for seed generation entirely.

\vspace{-1mm}
\subsection{Scoring Sets of Windows} \label{sec:scoring}

Recall that we evaluate sets of candidate windows using a scoring function. The function used is given in Algorithm \ref{algo:computeScore}. This returns the value of the objective function (Eq \ref{eq:concreteObjective}), with three alterations:
\begin{itemize}
\itemsep0em
\item We set the probabilities in the ``event'' model $\theta_1$ using the blurred windows.
\item We disallow features for which $\theta_1 < .5$, which is the minimum value that prevents the learning of two or more unrelated but frequent sets of features. This resembles a soft ``intersection'' operation and can be seen as a prior $p(\mathcal{F}|\theta_1)$.
\item We subtract the log odds of the windows being generated by noise or a ``rival'' event exemplified by the best candidate excluded. See \cite{extractWebsite} for a probabalistic interpretation of this operation.
\end{itemize}

\algnewcommand{\COMMENTT}[2][.5\linewidth]{\leavevmode\hfill\makebox[#1][l]{\hphantom{aa} // ~#2}}

\begin{algorithm}[h]
\caption{computeScore($\mathcal{I}, nextIdx$)}
\label{algo:computeScore}
\begin{algorithmic}[1]

\State {$ c \leftarrow \sum_{i \in \mathcal{I}} x_i $} \COMMENTT {feature counts}
\State {$ \widetilde{c} \leftarrow \sum_{i \in \mathcal{I}} \widetilde{x}_i $} \COMMENTT {blurred feature counts}
\State {$ \theta_1 \leftarrow \widetilde{c} \text { / } |\mathcal{I}| $} \COMMENTT {feature probabilities}
\State {$ \Delta \leftarrow \log{\theta_1} - \log{\theta_0} $} \COMMENTT {feature weights}
\State {$ \mathcal{F} \leftarrow \{j \text{ $|$ } \Delta_j > 0 \land \theta_{1j} > .5\} $\hphantom{ll}//\hphantom{h.}optimal features }
\State {$ w \leftarrow \langle \Delta_j \text{ if } j \in \mathcal{F} \text{ else } 0 \rangle $} \COMMENTT {feature weight vector}

\State {$ odds_{event} \leftarrow w \cdot c $}
\State {$ odds_{next} \leftarrow w \cdot x_{nextIdx} \cdot |\mathcal{I}| $}
\State {$ odds_{noise} \leftarrow \sum_{j \in \mathcal{F}} w_j \cdot E[\widetilde{\Phi}] \cdot |\mathcal{I}| $}

\RETURN {$ odds_{event} - \max(odds_{noise}, odds_{next}) $}


\end{algorithmic}
\end{algorithm}

In lines 1-3, we compute the counts of each feature and construct $\theta_1$ based on the data in $\mathcal{I}$. In lines 4-5, we determine the optimal set of features to include assuming irrelevant features are distributed according to $\theta_0$. In line 6, we construct a set of weights $w$ for the features. These weights are 0 for features that are not in $\mathcal{F}$ and equal to the difference between $\log(\theta_1)$ and $\log(\theta_0)$ for those that are. The introduction of $w$ is merely a convenience so that $w \cdot c = \sum_{j \in \mathcal{F}} c_j(\log(\theta_{1j}) - \log(\theta_{0j}))$, the original objective function. Line 7 computes the value of this objective, which can be seen as the increase in log likelihood from generating the ones in $\mathcal{I}$ using $\theta_1$ instead of $\theta_0$. Line 8 computes this increase in odds for the next window excluded, instead of for the supposed instances. Line 9 computes the increase in odds for an average ``noise'' window. 

The returned score corresponds to the log odds of a set of instances being generated by an event model versus either random noise or another event exemplified by the best candidate excluded. See \cite{extractWebsite} for a more detailed analysis.

\vspace{-1mm}
\subsection{Recovering Instance Bounds} \label{sec:instanceBounds}
Given an estimated set of instance-containing window positions $\mathcal{I}$, we compute $\mathcal{R}$ by discarding columns in the windows that are no more similar than chance.

Let $V$ be the feature weights $w$ in the above algorithm associated with $\mathcal{I}$ and reshaped to match the $J \times M_{max}$ shape of the window, where $J$ is the number of rows in $\Phi$. We sum the entries in each column of $V$ to produce a set of column scores, and subtract from each score the number of ones that would be expected by chance. This number is equal to $J \cdot E[\widetilde{\Phi}] ^{|\mathcal{I}|-1}$. We then extract the maximum subarray of the scores to find the start and end offsets of the ``event'' within $v$. We add these offsets to the indices in $\mathcal{I}$ to get $\mathcal{R}$.
This scheme is simple to implement, but does not guarantee optimal offsets.

\vspace{-1mm}
\subsection{Runtime Complexity} \label{sec:complexity}

We state the following without proof. The derivations are available at \cite{extractWebsite}.


\vspace*{-1mm}
\newtheorem{theorem}{Theorem}
\newtheorem{lemma}{Lemma}
\begin{lemma} Computing the structure scores for all subsequences requires $O(DM_{max}N)$ time.
\end{lemma}

\vspace*{-3mm}
\begin{lemma} Constructing the feature matrix requires \\ $O(DM_{max}N\log(N))$ time.
\end{lemma}

\vspace*{-3mm}
\begin{lemma} Optimizing the objective given the feature matrix and seeds requires $O(DM_{max}\log(M_{max})N\log(N))$ time.
\end{lemma}
Since these steps are sequential, the total running time of our algorithm is $O(DM_{max}\log(M_{max})N\log(N))$.

\section{Results} \label{sec:results}

We implemented our algorithm, along with baselines from the literature \cite{moen, plmd}, using SciPy \cite{scipy}. For the baselines, we JIT-compiled the inner loops using Numba \cite{numba}. All code and raw results are publicly available at \cite{extractWebsite}. Our full algorithm, including feature matrix construction, is under 300 lines of code. To our knowledge, our experiments use more ground truth event instances than any similar work.

\subsection{Datasets}

\begin{figure}[h]
\begin{center}
	\includegraphics[width=\linewidth]{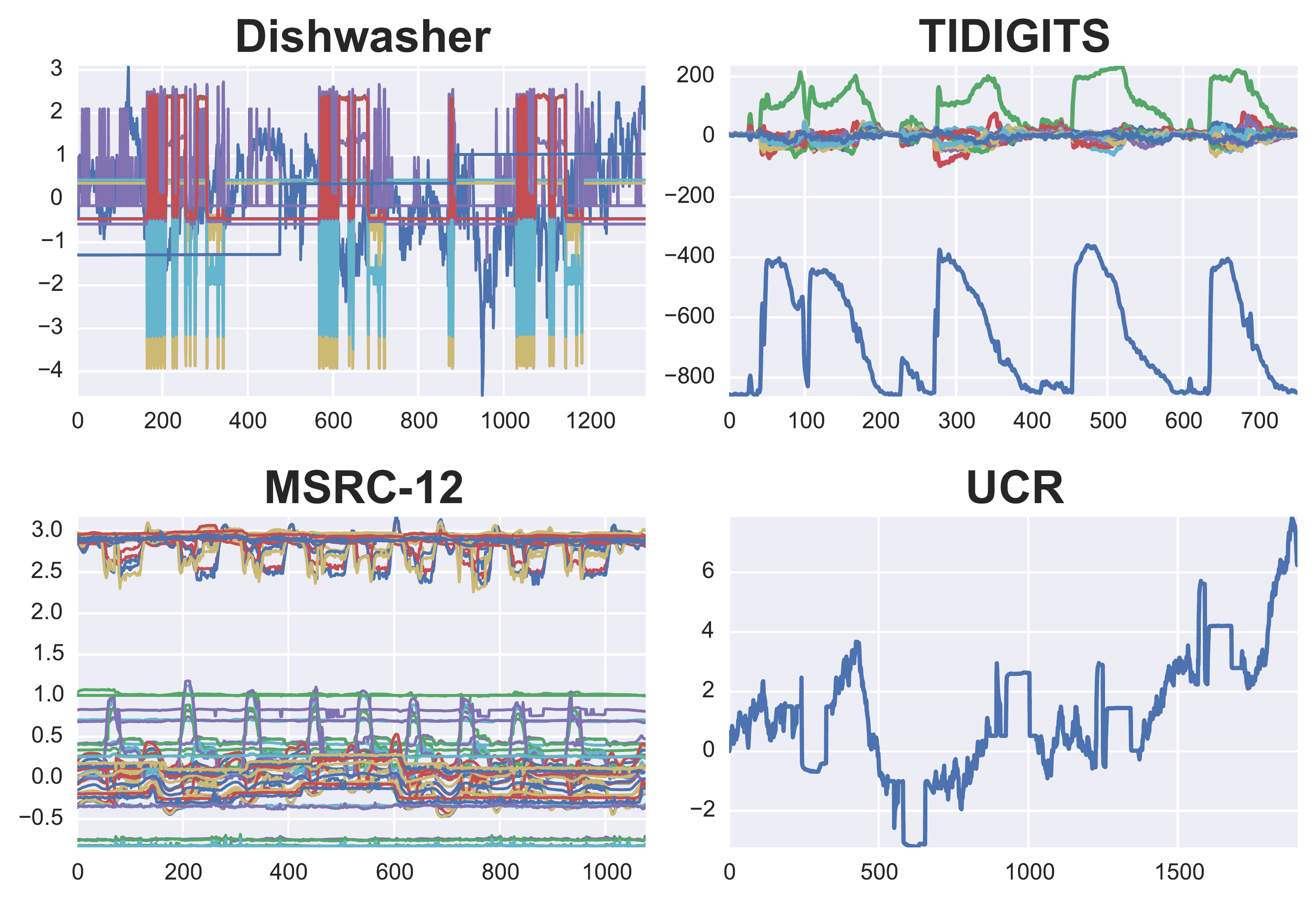}
	\caption{Example time series from each of the datasets used.}
	\label{fig:datasets}
\end{center}
\end{figure}

We used the following datasets (Fig \ref{fig:datasets}), selected on the basis that they were both publicly available and contained repeated instances of some ground truth event, such as a repeated gesture or spoken word.

Some of these events could be isolated with simpler techniques than those considered here---e.g., an appropriately-tuned edge detector could find many of the instances in the TIDIGITS time series. However, \textit{the goal of our work is to find events \textbf{without} requiring users to design features and algorithms for each domain or event of interest}. Thus, we deliberately refrain from exploiting dataset-specific features or prior knowledge. Moreover, such knowledge is rarely sufficient to solve the problem---even when one knows that events are periodic, contain peaks, etc., isolating their starts and ends programmatically is still challenging. 

To aid reproducibility, we supplement the source code with full descriptions of our preprocessing, random seeds, etc., at \cite{extractWebsite}, and omit the details here for brevity.

\subsubsection{MSRC-12} \label{sec:msrc}
The MSRC-12 dataset \cite{msrc12} consists of (x,y,z) human joint positions captured by a Microsoft Kinect while subjects repeatedly performed specific motions. Each of the 594 time series in the dataset is 80 dimensional and contains 8-12 event instances.

Each instance is labeled with a single marked time step, rather than with its boundaries, so we use the number of marks in each time series as ground truth. That is, if there are $k$ marks, we treat the first $k$ regions returned as correct. This is a less stringent criterion than on other datasets, but favors existing algorithms insofar as they often fail to identify exact event boundaries.

\subsubsection{TIDIGITS}
The TIDIGITS dataset \cite{tidigits} is a large collection of human utterances of decimal digits. We use a subset of the data consisting of all recordings containing only one type of digit (e.g., only ``9''s). We randomly concatenated sets of 5-8 of these recordings to form 1604 longer recordings in which multiple speakers utter the same word. As is standard practice \cite{minnenSubseqDensity}, we represented the resulting audio using Mel-Frequency Cepstral Coefficients (MFCCs) \cite{librosa}, rather than as the raw speech signal. Unlike in the other datasets, little background noise and few transient phenomena are present to elicit false positives; however, the need to generalize across speakers and rates of speech makes avoiding false negatives difficult. 

\subsubsection{Dishwasher}
The Dishwasher dataset \cite{ampds} consists of energy consumption and related electricity metrics at a power meter connected to a residential dishwasher. It contains twelve variables and two years worth of data sampled once per minute, for a total of 12.6 million data points.

We manually plotted, annotated, and verified event instances across all 1 million+ of its samples.\footnote{See our supporting website \cite{extractWebsite} for details.}

Because this data is 100x longer than what the comparison algorithms can process in a day \cite{plmd}, we followed much the same procedure as for the TIDIGITS dataset. Namely, we extracted sets of 5-8 event instances, along with the data around them (sometimes containing other transient phenomena), and concatenated them to form shorter time series.

\subsubsection{UCR}
Following \cite{plmd}, we constructed synthetic datasets by planting examples from the UCR Time Series Archive \cite{ucrTimeSeries} in random walks. We took examples from the 20 smallest datasets (before the 2015 update), as measured by the lengths of their examples. For each dataset, we created 50 time series, each containing five examples of one class. This yields 1000 time series and 5000 instances.


\subsection{Evaluation Measures}
Let $\mathcal{R}$ be the ground truth set of instance regions and let $\mathcal{\hat{R}}$ be the set of regions returned by the algorithm being evaluated. Further let $r_1 = (a_1, b_1)$ and $r_2 = (a_2, b_2)$ be two regions.
\begin{Definition}{$IOU(r_1, r_2)$. The \textbf{Intersection-Over-Union (IOU)} of $r_1$ and $r_2$ is given by $|r_1 \cap r_2| / |r_1 \cup r_2|$, where $r_1$ and $r_2$ are treated as intervals.}
\end{Definition}

\vspace{-3mm}
\begin{Definition}{$Match(r_1, r_2, \tau)$. $r_1$ and $r_2$ are said to \textbf{Match} at a threshold of $\tau$ iff $IOU(r_1, r_2) \ge \tau$.
}
\end{Definition}
\vspace{-3mm}
\begin{Definition}{MatchCount$(\mathcal{\hat{R}}, \mathcal{R}, \tau)$. The \textbf{MatchCount} of $\mathcal{\hat{R}}$ given $\mathcal{R}$ and $\tau$ is the greatest number of matches at threshold $\tau$ that can be produced by pairing regions in $\mathcal{\hat{R}}$ with regions in $\mathcal{R}$ such that no region in either set is present in more than one pair.\footnote{Since regions (and their possible matches) are ordered in time, this can be computed greedily after sorting $\mathcal{\hat{R}}$ and $\mathcal{R}$.}
}
\end{Definition}
\begin{Definition}{\textbf{Precision, Recall}, and \textbf{F1 Score}.
}
\end{Definition}
\vspace*{-6mm}
\begin{align}
	Precision(\mathcal{\hat{R}}, \mathcal{R}, \tau)
		&= MatchCount(\mathcal{\hat{R}}, \mathcal{R}, \tau) / |\mathcal{\hat{R}}| \\
	Recall(\mathcal{\hat{R}}, \mathcal{R}, \tau)
		&= MatchCount(\mathcal{\hat{R}}, \mathcal{R}, \tau) / |\mathcal{R}| \\
	F1(\mathcal{\hat{R}}, \mathcal{R}, \tau)
		&= \frac{2 \cdot Precision \cdot Recall}{Precision + Recall}
\end{align}


\subsection{Comparison Algorithms}

While none of the techniques we reviewed both seek to solve our problem and operate under assumptions as relaxed as ours, we found that two existing algorithms solving the univariate version of the problem could be generalized to the multivariate case:
\begin{enumerate}
\itemsep.1em
	\item Finding the closest pair of subsequences under the z-normalized Euclidean distance, and returning as instances all subsequences within some threshold distance of this pair \cite{moen, minnenSubdim}. In our case, distance is defined as the sum of the distances for each dimension, normalized individually. We find the closest pair efficiently using the MK algorithm \cite{mk} plus the length-pruning technique of Mueen \cite{moen}. We determine the distance threshold using Minnen's algorithm \cite{minnenNeighborhood}. We call this algorithm \textit{Dist}.
	\item The single-motif-finding subroutine of \cite{plmd}, with distances and description lengths summed over dimensions. This amounts to closest-pair motif discovery to find seeds, candidate generation based on Euclidean distance to these seeds, and instance selection using a Minimum Description Length (MDL) criterion.\footnote{In the case of a single event type, this maximizes the same objective as \cite{epenthesis}, but requires fewer closest-pair searches. We therefore compare only to the subroutine of \cite{plmd}.} We call this algorithm \textit{MDL}.
\end{enumerate}

In both cases, we consider versions of the algorithms that carry out searches at lengths from $M_{min}$ to $M_{max}$ and use the best result from any length. This means lowest distance in the former case and lowest description length in the latter. In other words, we give them the prior knowledge that there is exactly one type of event to be found, as well as its approximate length. This replaces the heuristics for determining the number of event classes described in the original papers. 

We tried many variations of these two algorithms regarding threshold function, description length computation, and other properties, and use the above approaches because they worked the best.

\subsection{Instance Discovery Accuracy}
\begin{figure}[h]
\begin{center}
	\includegraphics[width=\linewidth]{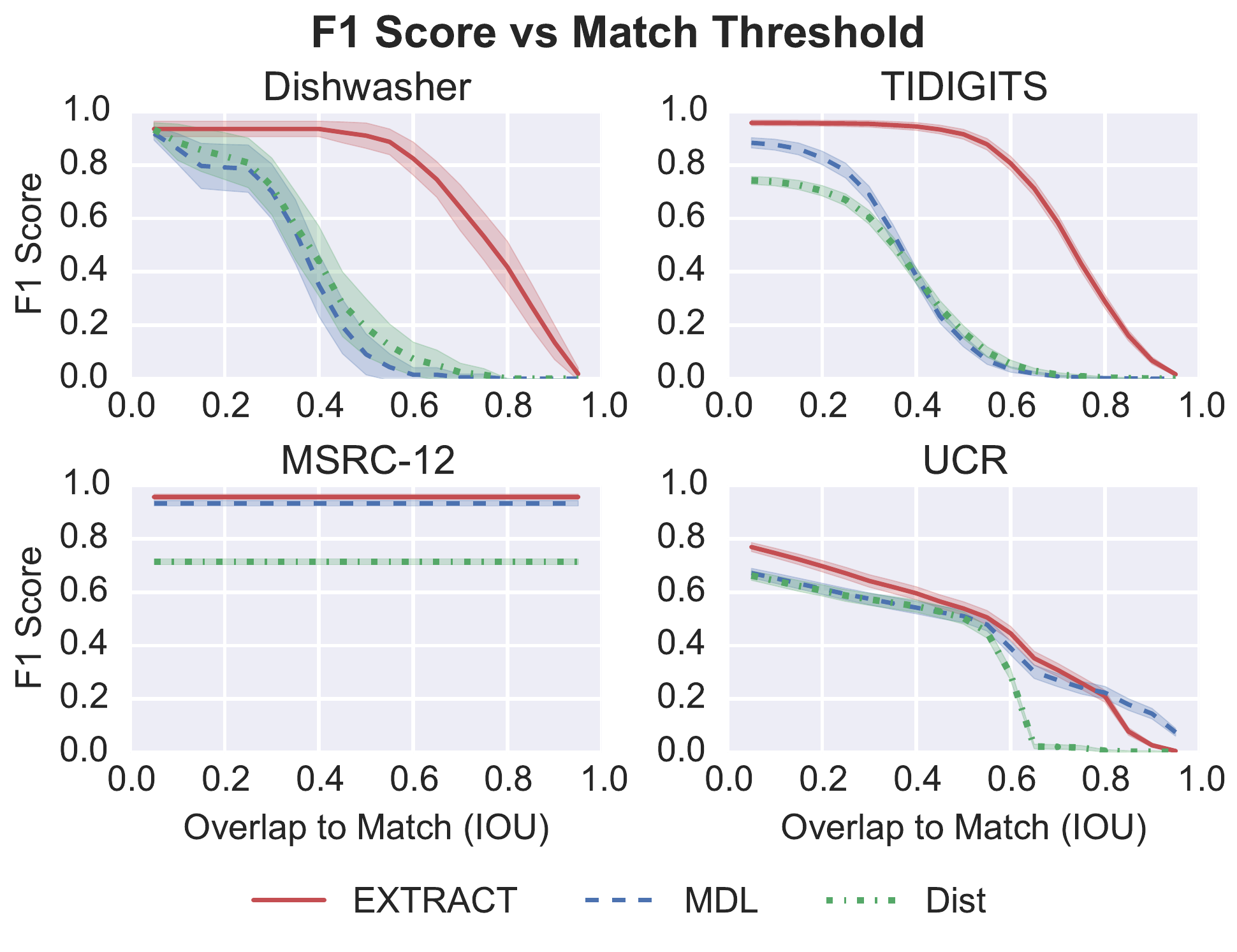}
	\vspace*{-6mm}
	\caption{The proposed algorithm is more accurate for virtually all ``match'' thresholds on all datasets. Shading corresponds to 95\% confidence intervals.}
	\label{fig:accuracy}
\end{center}
\end{figure}

The core problem addressed by our work is the robust location of multiple event instances within a time series known to contain a small number of them. To assess our effectiveness in solving this problem, we evaluated the F1 score on each of the four datasets, varying the threshold $\tau$ for how much ground truth and reported instances needed to overlap in order to count as matching. In all cases, $M_{min}$ and $M_{max}$ were set to $1/20$ and $1/10$ of the time series length. As shown in Figure~\ref{fig:accuracy}, we outperform the comparison algorithms for virtually all match thresholds on all datasets.
\vspace{-1mm}

\vspace{1mm}
Note that MSRC-12 values are constant because instance boundaries are not defined in this dataset (see Section \ref{sec:msrc}). Further, the dataset on which we perform the closest to the comparisons (UCR) is synthetic, univariate, and only contains instances that are the same length. These last two attributes are what \textit{Dist} and \textit{MDL} were designed for, so the similar F1 scores suggest that \textit{EXTRACT}'s superiority on other datasets is due to its robustness to violation of these conditions. Visual examination of the errors on this dataset suggests that all algorithms have only modest accuracy because there are often regions of random walk data that are more similar in shape to one another than the instances are.

The dropoffs in Figure~\ref{fig:accuracy} at particular IOU thresholds indicate the typical amount of overlap between reported and true instances. E.g., the fact that existing algorithms abruptly decrease in F1 score on the TIDIGITS dataset at a threshold near 0.3 suggests that many of their reported instances only overlap this much with the true instances.

\vspace{-.5mm}
Our accuracy on real data is not only superior to the comparisons, but also high in absolute terms (Table~1). Suppose that we consider IOU thresholds of 0.25 or 0.5 to be ``correct'' for our application. The former might correspond to detecting a portion of a gesture, and the latter might correspond to detecting most of it, with a bit of extra data at one end. At each of these thresholds, our algorithm discovers event instances with an F1 score of over $.9$ on real data.
\vspace{5mm}
\begin{table}[h]
\setlength\tabcolsep{4pt} 
\centering
\caption*{Table 1: EXTRACT F1 Scores are High in Absolute Terms}
\label{tbl:f1}
\begin{tabularx}{\linewidth}{@{\hskip-1.5pt}llll|lll}
\hline
& \multicolumn{3}{c}{Overlap $\ge .25$} & \multicolumn{3}{c}{Overlap $\ge .5$} \\
& Ours &   MDL & Dist & Ours &  MDL & Dist \\
\hline
Dishwasher & \textbf{0.935} & 0.786 &  0.808 & \textbf{0.910} & 0.091 &  0.191 \\
TIDIGITS   & \textbf{0.955} & 0.779 &  0.670 & \textbf{0.915} & 0.140 &  0.174 \\
MSRC-12    & \textbf{0.947} & 0.943 &  0.714 & \textbf{0.947} & 0.943 &  0.714 \\
UCR        & \textbf{0.671} & 0.593 &  0.587 & \textbf{0.539} & 0.510 &  0.504 \\
\hline
\end{tabularx}
\end{table}
\vspace{-1mm}

\begin{figure}[t]
\begin{center}
	\includegraphics[width=\linewidth]{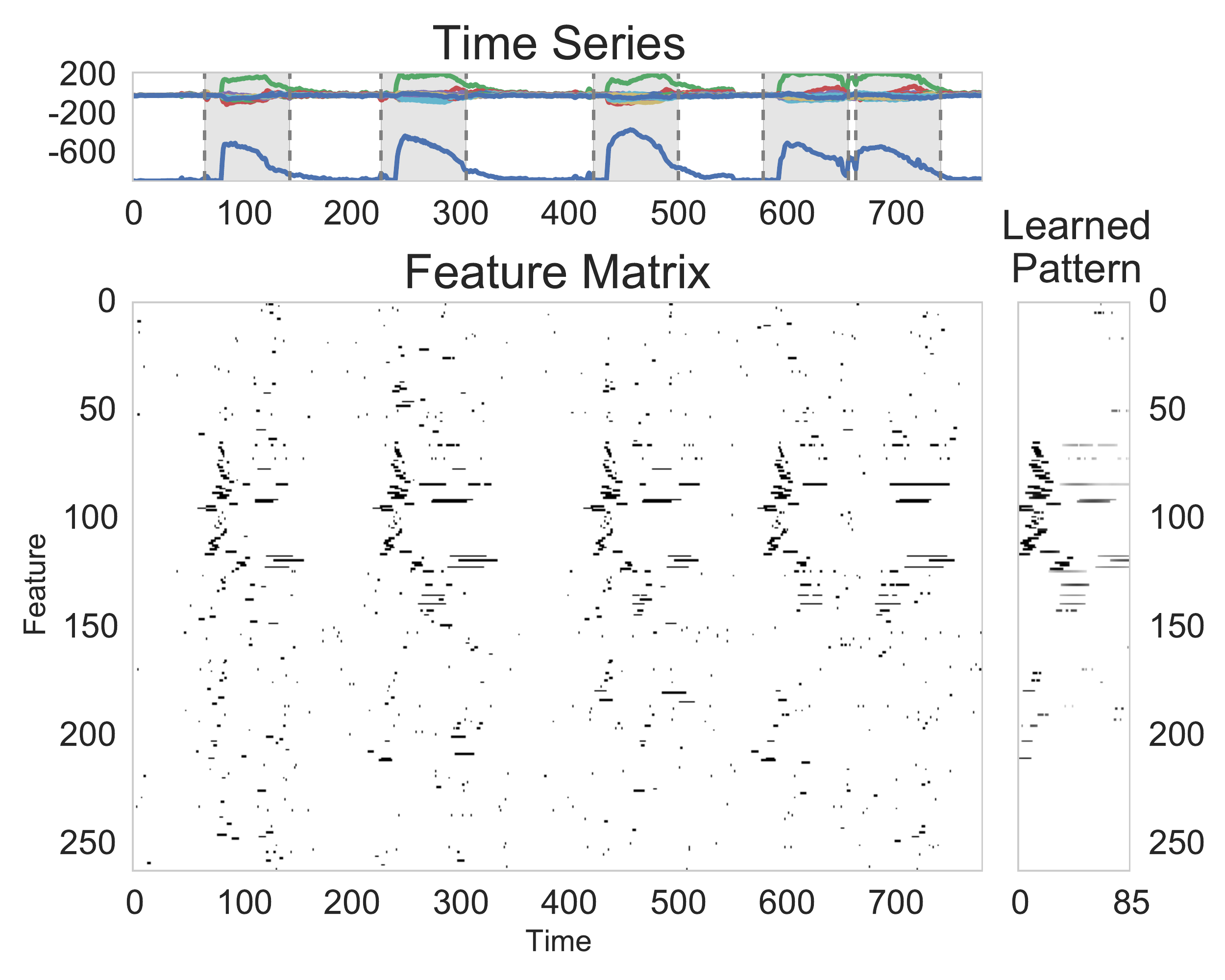}
	\vspace*{-6mm}
	\caption{\textit{Top}) Original time series, with instances inferred by EXTRACT in gray. \textit{Bottom}) The feature matrix $\Phi$. \textit{Right}) The learned feature weights. These resemble a ``blurred'' version of the features that occur when the word is spoken.}
	\label{fig:algoOutput}
\end{center}
\end{figure}

An example of our algorithm's output on the TIDIGITS dataset is shown in Figure~\ref{fig:algoOutput}. The regions returned (shaded) closely bracket the individual utterances of the digit ``0.'' The ``Learned Pattern'' is the feature weights $V$ from Section~\ref{sec:instanceBounds}, which are the increases in log probability of each element being 1 when the window is an event instance.

\subsection{Speed}
In addition to being accurate on both real and synthetic data, our algorithm is also fast. To assess performance, we recorded the time it and the comparisons took to run on increasingly long sections of random walk data and the raw Dishwasher data.

\begin{figure}[h]
\begin{center}
	\includegraphics[width=\linewidth]{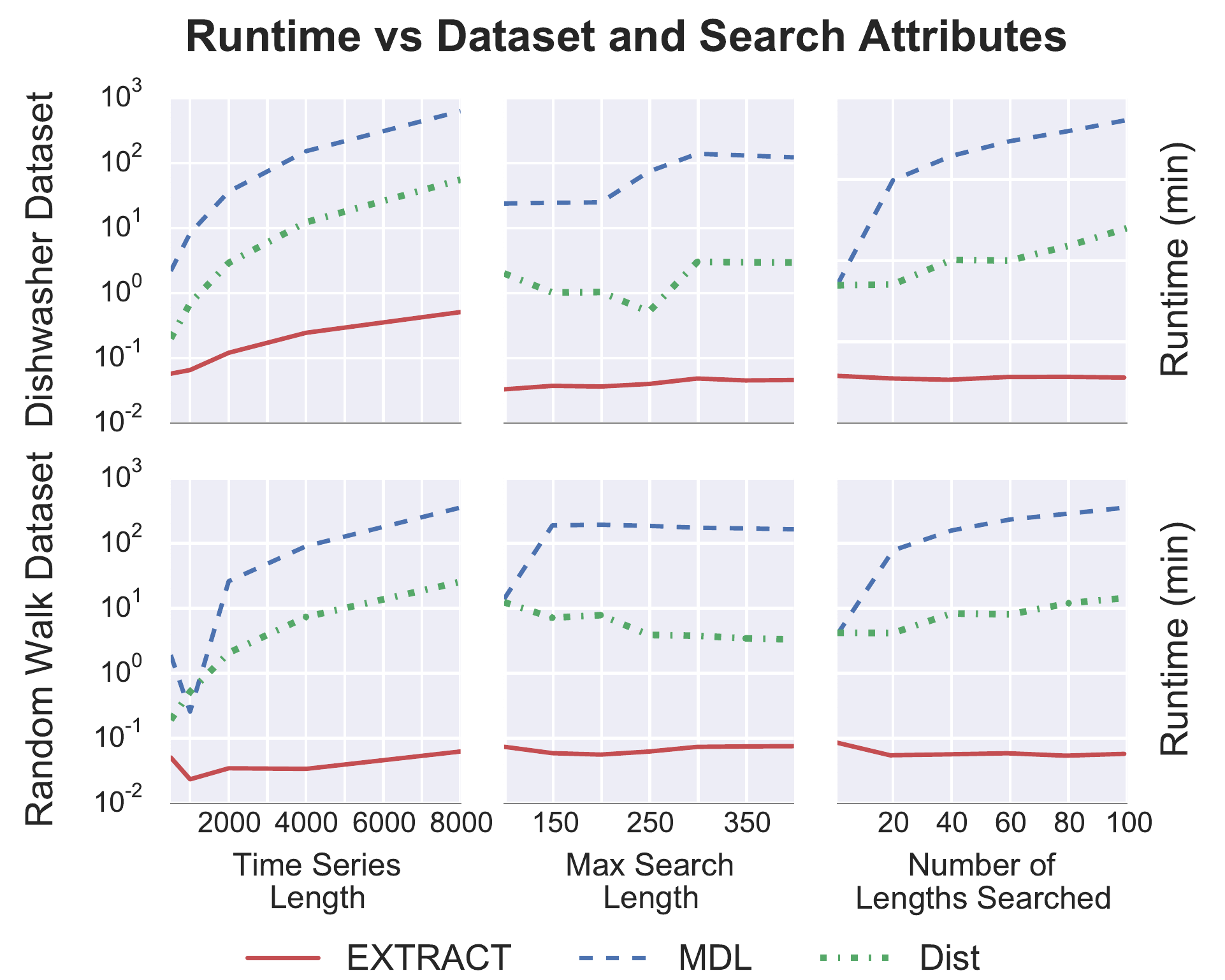}
	\caption{The proposed algorithm is one to two orders of magnitude faster than comparisons.}
	\label{fig:scalability}
\end{center}
\end{figure}

In the first column of Fig \ref{fig:scalability}, we vary only the length of the time series ($N$) and keep $M_{min}$ and $M_{max}$ fixed at 100 and 150. In the second column, we hold $N$ constant at 5000 and vary $M_{max}$, with $M_{min}$ fixed at $M_{max} - 50$ so that the number of lengths searched is constant. In the third column, we fix $N$ at 5000 and set ($M_{min}$, $M_{max}$) to $(150,150),(140,160),$\ldots$,(100,200)$.

Our algorithm is at least an order of magnitude faster in virtually all experimental conditions. Further, it shows little or no increase in runtime as $M_{min}$ and $M_{max}$ are varied and increases only slowly with $N$. This is in line with what would be expected given our computational complexity, except with even less dependence on $M_{max}$. This deviation is because $D\log(M_{max})\log(N)$ is an upper bound on the number of features used---the actual number is lower since shapes that only occur once are discarded. This is also why our algorithm is faster on the Random Walk dataset than the Dishwasher dataset; random walks have few repeating shapes, so our feature matrix has few rows.

Both \textit{Dist} and \textit{MDL} sometimes plateau in runtime thanks to their early-abandoning techniques. \textit{Dist} even decreases because the lower bound it employs \cite{moen} to prune similarity comparisons is tighter for longer time series. Since they are $O(N^2)$, they are also helped by the  decrease in the number of windows to check as the maximum window length $M_{max}$ increases. This decrease benefits EXTRACT as well, but to a lesser extent since it is subquadratic.

As with accuracy, our technique is fast not only relative to comparisons, but also in absolute terms---we are able to run the above experiments in minutes and search each time series in seconds (even with our simple Python implementation). Since these time series reflect phenomena spanning many seconds or hours, this means that our algorithm could be run in real time in many settings.

\vspace{-1mm}
\section{Discussion and Conclusion} \label{sec:conclusion}

We have described an algorithm to efficiently and accurately locate instances of an event within a multivariate time series given virtually no prior information about the nature of this event. In particular, we assume no knowledge of how many times the event has occurred, what features distinguish it, or which variables it affects. Using a diverse group of publicly available datasets, we showed that this technique is fast and accurate both in absolute terms and compared to existing algorithms, despite its limited assumptions.

Moreover, while this work has focused on a feature matrix reflecting the presence of particular shapes in the data, our technique could be applied even when signals are not described well by shapes---our learning algorithm requires only a sparse feature matrix with entries between 0 and 1. In particular, one could one-hot encode categorical variables such as ``day of week'' or ``user gender'' and add these features with no change to the algorithm. We consider this adaptability a major strength of our approach, since mixed real and categorical variables are common in many domains.

In short, by applying our technique to low-level signals of various kinds, one can isolate segments of data produced by high-level events as diverse as spoken words, human actions, and household appliance usage.

\section{Acknowledgements}
This work was supported by NSF Grant 020772-00001.



\bibliographystyle{IEEEtran}
\bibliography{doc}

\end{document}